# Automatic Plant Image Identification of Vietnamese species using Deep Learning Models

Nguyen Van Hieu[1], Ngo Le Huy Hien[2]

[1]Faculty of Information Technology, The University of Danang - University of Science and Technology
[2]Faculty of Computer Science and Engineering, The University of Danang - VN-UK Institute for Research and Executive Education

***Abstract*** *It is complicated to distinguish among thousands of plant species in the natural ecosystem, and many efforts have been investigated to address the issue. In Vietnam, the task of identifying one from 12,000 species requires specialized experts in flora management, with thorough training skills and in-depth knowledge. Therefore, with the advance of machine learning, automatic plant identification systems have been proposed to benefit various stakeholders, including botanists, pharmaceutical laboratories, taxonomists, forestry services, and organizations. The concept has fueled an interest in research and application from global researchers and engineers in both fields of machine learning and computer vision. In this paper, the Vietnamese plant image dataset was collected from an online encyclopedia of Vietnamese organisms, together with the Encyclopedia of Life, to generate a total of 28,046 environmental images of 109 plant species in Vietnam. A comparative evaluation of four deep convolutional feature extraction models, which are MobileNetV2, VGG16, ResnetV2, and Inception Resnet V2, is presented. Those models have been tested on the Support Vector Machine (SVM) classifier to experiment with the purpose of plant image identification. The proposed models achieve promising recognition rates, and MobilenetV2 attained the highest with 83.9%. This result demonstrates that machine learning models are potential for plant species identification in the natural environment, and future works need to examine proposing higher accuracy systems on a larger dataset to meet the current application demand.*

## I. INTRODUCTION

As the ecosystem bears 391,000 plant species, it is crucial to research and classify them correctly in order to manage and protect the natural flora. However, plant identification requires expert botanists who possess inherent knowledge and in-depth skills in botany and plant systematics, leading to time-consuming and laborious. Therefore, global researchers have conducted various studies to assist the plant recognition process by applying image classification systems. By analyzing physical characteristics and identifying plant species through photographs, these systems draw tremendous attention from many researchers and engineers in the field of computer vision and machine learning. Many works have been implemented using a varying number of methods and models of deep learning advances. Sophisticated models have also been proposed so far to construct machine-learning aided plant identification systems. Following this trend, many research authors have attempted to improve the performance of automatic recognition systems.

Over the last decade, there are numerous studies that have been implemented to construct automatic identification systems of plants by many authors and attained promising outcomes. In early periods, global researchers have used leaves as a standard physical characteristic to distinguish among different species, using features of texture, shape, and color [1]-[4]. Nam et al. [5]-[6] have proposed a shape-based search method in their plant identification system. By utilizing not only the leaves outline but also their vein data, they attempted to enhance the accuracy using the Minimum Perimeter Polygons (MPP) algorithm and Extended Curvature Scale Space (CSS). Training on a dataset that consists of 1,032 plant leaf images, a weighted graph was presented and deliver a positive recognition rate. Another tree identification algorithm has been proposed by Aakif et al. [7] by conducting three steps, from preprocessing to extraction, and sorting. They used Artificial Neural Network (ANN) to classify the leaf morphological characteristics, Fourier descriptions, and shape. The algorithm has attained an accuracy of more than 96% on 817 different leaf samples from 14 fruit trees. In [8], Wang-Su and his colleague have delivered a new method to classify leaves by using the Convolutional Neural Network (CNN) model, and the other two models by using GoogleNet to adjust the network depth. These models achieved greater than 94% in the highest case, even when the testing leaf was 30% damaged.

It is perceivable that most of the mentioned studies have concentrated on using hand-crafted image features for plant identification; however, they reach a common limitation by this approach. Noise and background are the factors that affect low-level image representation on most of these hand-crafted features. Carranza-Rojas and Mata-Montero (2016) have created a work to prove the affection of the noise and background, in which they created two datasets: a clean one and a noisy one [9]. The result has figured that the clean dataset outperformed the noisy dataset by at least 7.3%. This result claims that images manually processed in a laboratory and then classified produces a higher level of satisfactory accuracy compared to images that are taken directly on smartphones. Therefore, to apply in practical uses, it is challenging to utilize the very clean input images without any background that the mentioned authors applied. For this reason, to reach higher efficiency of recognition and retrieval images of plants in the real world, it requires designing a high-level representation image with less affecting by the environment. Many researchers have theoretically and practically focused on this application [10]-[11]. Applying twelve morphological features of a dataset of natural images from 20 species, Du et al. (2007) [12] achieved 93% accuracy in their proposed system by using the k-nearest neighbor classifier. In 2009, they reached 92.35 on their recognition rate on a larger dataset of 2000 environmental images from 20 different leaves [13]. A new approach has been proposed by them in 2013, which bases on fractal dimension features of shape and vein patterns of leaves [14]. They finally attained 87.1% accuracy in their plant identification system with k-nearest neighbors using 20 features. In an effort to create a plant image dataset in the natural environment, Yu et al. [15] have taken 10,000 images by mobile phones of 100 ornamental plant species that grow around the Beijing Forestry University campus to build the BJFU100 dataset. An uncontrolled plant identification system was created on a 26-layer deep learning model, using eight residual building blocks. They finally reached 91.78% in the recognition rate on this natural image dataset of plant species.

Inspired by the fast-improving plant identification models and overcoming the aforementioned challenge, a natural plant image dataset was collected from a national online encyclopedia of Vietnamese creatures together with the Encyclopedia of Life (EoL), containing 28,046 environmental images of 109 plant species in Vietnam. In this paper, four deep learning models of image feature extraction were applied to implement with the Support Vector Machine (SVM) classifier in the plant image recognition system. In particular, MobileNetV2, VGG16, ResnetV2, and Inception Resnet V2 have been experimented respectively to deliver a comparative evaluation in the end. The highest recognition rate that achieved was 83.9% on the MobilenetV2 model. This result is promising for the identification task of natural images from 109 plant species. The evaluation also delivers a solution for choosing a suitable feature extraction model for future plant image recognition systems.

## II. PLANT IMAGE RECOGNITION

### A. *Deep Convolutional Feature Extraction Models*

#### a) *MobileNetV2*

MobileNet is a backbone model for feature extraction that has been widely used theoretically and practically. Moreover, it has a state-of-the-art performance for object detection and semantic segmentation. In the MobileNet model [16], the Depthwise Separable Convolution can significantly reduce the model size and complexity of the network, which is applicable to mobile phones and low computational power devices. Compared to MobileNetV1, MobileNetV2 has a better module with an inverted residual structure and without non-linearities in narrow layers.

Using different width multipliers and input resolutions, MobileNetV2 outperforms MobileNetV1, while it has a comparable computational cost and model size. Moreover, MobileNetV2, which has a width multiplier of 1.4, performs better with faster inference time than ShuffleNet (×2) and NASNet [16]. Therefore, MobileNetV2 is an outstanding feature extraction model for object detection and segmentation.

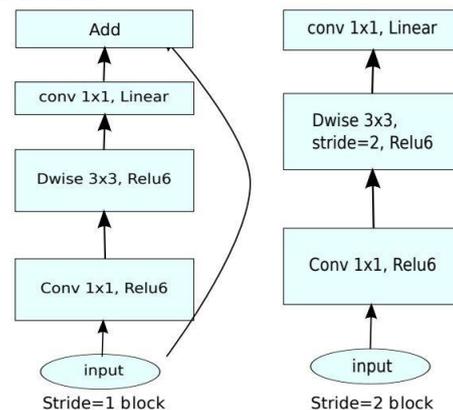

Fig. 1. MobileNetV2 Convolutional Blocks [17].

In the MobileNetV2 architecture [17], as indicated in Fig. 1, there are two types of blocks: a residual block with a stride of 1, and another one with a stride of 2 for downsizing. Both types of blocks have three layers, including a 1×1 convolution with ReLU6 in the first layer, the second layer containing depth-wise convolution, and another 1×1 convolution without non-linearity in the third layer. The input, operator, and output of each layer are shown in Fig. 2.

| Input | Operator | Output |
|---|---|---|
| $h \times w \times k$ | 1x1 conv2d , ReLU6 | $h \times w \times (tk)$ |
| $h \times w \times tk$ | 3x3 dwise s=s, ReLU6 | $\frac{h}{s} \times \frac{w}{s} \times (tk)$ |
| $\frac{h}{s} \times \frac{w}{s} \times tk$ | linear 1x1 conv2d | $\frac{h}{s} \times \frac{w}{s} \times k'$ |

Fig. 2. Three layers of the MobileNetV2 [16].

In Fig. 2, t=6 for all experiments, and t contributes as an expansion factor. If there are 64 channels in the input, the internal output would contain 64×t=64×6=384 channels. It is figured that if ReLU is reused, the deep networks just have a similar power of a linear classifier on a non-zero volume part of the output domain.

| Input | Operator | t | c | n | s |
|---|---|---|---|---|---|
| $224^2 \times 3$ | conv2d | - | 32 | 1 | 2 |
| $112^2 \times 32$ | bottleneck | 1 | 16 | 1 | 1 |
| $112^2 \times 16$ | bottleneck | 6 | 24 | 2 | 2 |
| $56^2 \times 24$ | bottleneck | 6 | 32 | 3 | 2 |
| $28^2 \times 32$ | bottleneck | 6 | 64 | 4 | 2 |
| $14^2 \times 64$ | bottleneck | 6 | 96 | 3 | 1 |
| $14^2 \times 96$ | bottleneck | 6 | 160 | 3 | 2 |
| $7^2 \times 160$ | bottleneck | 6 | 320 | 1 | 1 |
| $7^2 \times 320$ | conv2d 1x1 | - | 1280 | 1 | 1 |
| $7^2 \times 1280$ | avgpool 7x7 | - | - | 1 | - |
| $1 \times 1 \times 1280$ | conv2d 1x1 | - | k | - |   |

Fig. 3. The overall architecture of the MobileNetV2 [16].

Fig. 3. illustrates the overall architecture of the MobileNetV2, where c is the number of output channels, n is the repeating number, s is the stride, and 3×3 kernels using for spatial convolution. Ordinarily, the primary network, which has width multiplier 1, 224×224, uses 3.4 million parameters and has a computational cost of 300 million multiply-adds. While the model size varies between 1.7 million and 6.9 million parameters, the computational network cost only reaches less than 585 million MAdds.

*b) VGG16*

VGG16 is a CNN model improving the classification accuracy by adding depth, which is proposed by K. Simonyan and A. Zisserman from the University of Oxford [18]. VGG16 obtains 92.7% on top-5 accuracy on the ImageNet dataset which consists of more than 14 million images of 1000 classes. This architecture was one of the famous models submitted and won the ImageNet ILSVR competition 2014. It improves AlexNet by replacing its large kernel-sized filters with multiple 3×3 kernel-sized filters. Fig. 4 illustrates the VGG16 architecture, where a fixed size 224 x 224 RGB image is the input of the cov1 layer [19]. A stack of convolutional layers passes through this image by filters with a small receptive field: 3×3. 1×1 convolution filter is also used in one of the configurations to make a linear transformation for the input channels. Spatial pooling performs by five max-pooling layers after some convolutional layers. Max-pooling operates with stride 2 on a 2×2-pixel window. There are three fully-connected layers after the stack of convolutional layers, each of the first two has 4096 channels, and the third contains 1000 channels. The soft-max layer is the final layer. The fully connected layers configuration is the same in all networks, and all hidden layers include rectification (ReLU) non-linearity.

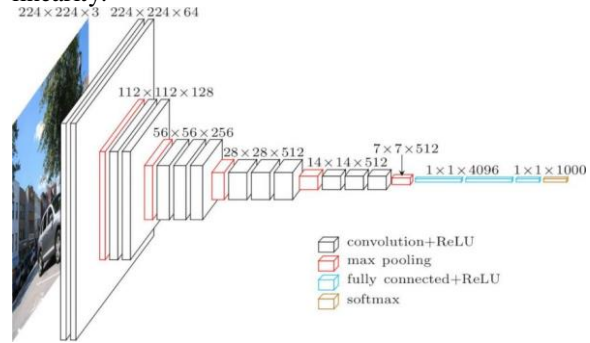

Fig. 4. The architecture of VGG16 [19].

*c) ResnetV2*

When training artificial neural networks using gradient-based learning and backpropagation, it encountered the vanishing gradient problem, leading to the network to stop training. Therefore, Residual Neural Networks, or commonly known as ResNets, are proposed to solve the problem. ResNets are neural networks that apply identity mapping, in which some layer inputs are passed directly to some other layers by a so-called "identity shortcut connection". Achieving 3.57% in the error rate, Resnet became the winner of ImageNet ILSVRC [20] and MS COCO [21] competitions in 2015. The left chart in Fig. 5 shows the original residual block, ResNet v1.

Residual Neural Networks (ResNets) [22] contain many stacked called "Residual Units", which has a general form of each unit as:

$$\mathbf{y}_l = h(\mathbf{x}_l) + A(\mathbf{x}_l, W_l), \quad \mathbf{x}_{l+1} = f(\mathbf{y}_l),$$

where $\mathbf{x}_l$ and $\mathbf{x}_{l+1}$ are input and output of the *l*-th unit, and A is a residual function. In [22], $h(\mathbf{x}_l) = \mathbf{x}_l$ is an identity mapping and *f* is a ReLU [23] function. The core idea is to find out the additive residual function A with regard to $h(\mathbf{x}_l)$, and using an identity mapping $h(\mathbf{x}_l) = \mathbf{x}_l$ as a key choice.

ResNet v2 has some prominent changes in the arrangement of the layers in the residual block, as illustrated in Fig. 5. They include: using Batch normalization. a stack of $1 \times 1 - 3 \times 3 - 1 \times 1$ BN-ReLU-Conv2D, and ReLU activation before 2D convolution [24].

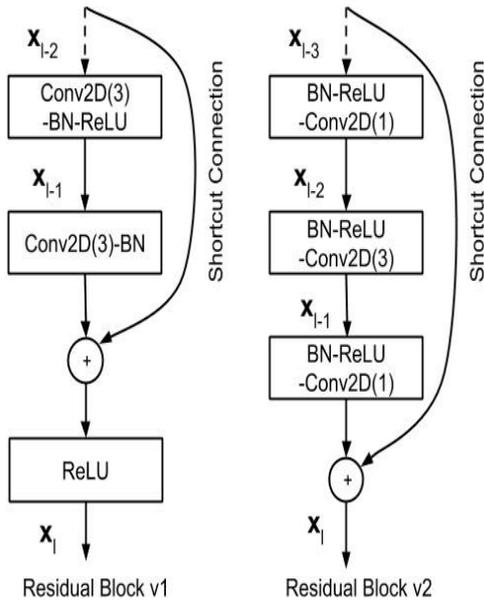

Fig. 5. A comparison between ResNet v1 and ResNet v2 on residual blocks.

*d) Inception Resnet V2*

Inception-ResNet-v2 is a CNN that achieves an outstanding accuracy on the ILSVRC image classification benchmark. Borrowing some ideas from Microsoft's ResNet papers [25][26], Inception-ResNet-v2 is a variation of the Inception V3 model. It merges together the concepts of Inception V3 and ResNet architectures [27]. In [28], Christian et al. explained all details of the Inception-ResNet-v2 model. Overall, residual connections have better performance and significant simplification to train in very deep neural networks by shortcuts in the model.

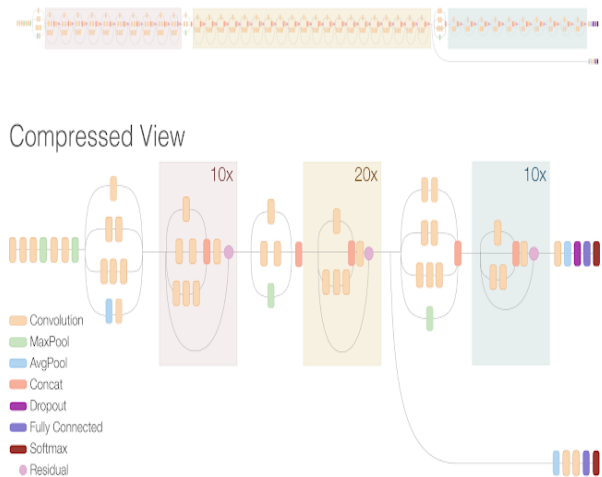

Fig. 6. Inception-ResNet-v2 schematic diagram [28].

At the top of Fig. 6, it indicates a full network expanded of the Inception-ResNet-v2, which is considerably deeper than Inception V3. The more accessible explained version of the same network is presented in the below figure, where repeated residual blocks are compressed. It is noticeable that the inception blocks have been simplified, including fewer parallel towers than Inception V3. In terms of accuracy, the Inception-ResNet-v2 performed better than previous state-of-the-art models, including Inception V3, ResNet 152, and ResNet V2 200.

*B. Classification Method*

In this research, the Support Vector Machine (SVM) was used in the plant image classification system. SVM is a supervised machine learning algorithm, which commonly used for classification rather than regression purposes. It is a state-of-the-art classifier and widely used in various classification applications of input samples [29, 30]. Let $\{(x_i, y_i)\}_{i=1}^{N}$ be a set of N training samples, where $x_i$ is the i[th] sample in the input space x, and $y_i \in \{+1, -1\}$ is the class of $x_i$ label. The decision function of SVM that classifies a new test sample $x$ can be represented as

$$f(z) = \text{sgn}\left(\sum_{i=1}^{N} \alpha_i y_i k(x_i, z) + b\right)$$

where $z$ is an unclassified sample, $\alpha_i$ is the Lagrange multiplier of a dual optimization problem that describes the separating hyperplane; $k(\cdot, \cdot)$ denotes the kernel function which should satisfy Mercer's condition; and $b$ is the hyperplane threshold parameter [31]. The training sample $x_i$ (with $\alpha_i > 0$) is called support vectors, and the SVM classifier finds the optimal hyperplane that maximizes the separating margin between two classes, as shown in Fig. 5 [32].

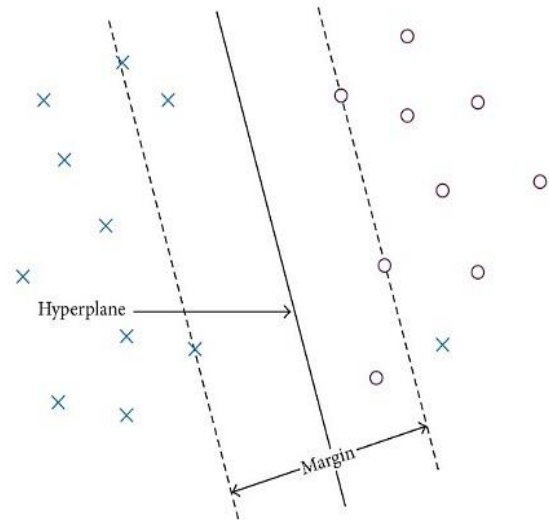

Fig. 7. Optimal hyperplane of SVM in non-separable cases [32].

## III. EXPERIMENTAL RESULTS

### A. Dataset Collection

The plant image dataset was collected from vncreatures.net [33], an online encyclopedia of Vietnamese organisms, and consists of thousands of plant species in Vietnam. The site includes detailed descriptions in Vietnamese, local names, scientific nomenclature, and natural images. Although its description data is quite complete, the number of illustrations for each species remains small, with only about 2 to 5 images. Such a limited number of images cannot guarantee the training efficiency of machine learning models; therefore, the project uses another data source to supplement the number of images.

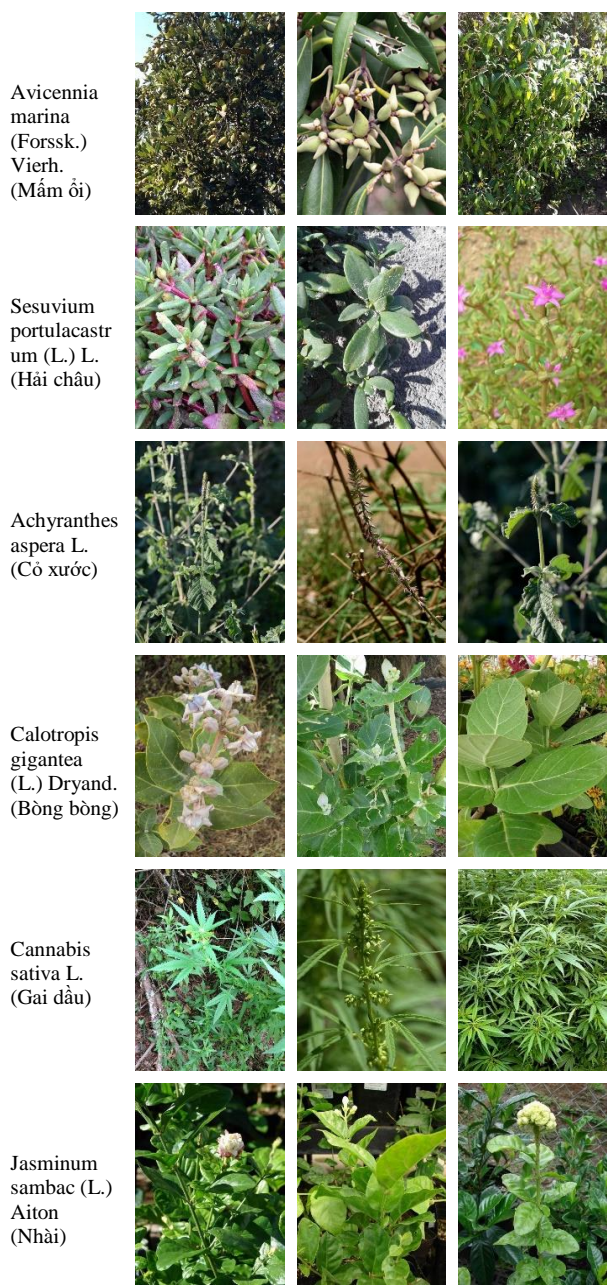

Fig. 8. Example images of the Vietnamese plant species training dataset. The left column is the class name, and the next columns are three sample images taken in the real-life environment.

Encyclopedia of Life (EoL) [34] is an online encyclopedia about living creatures on earth. This site aggregates reliable sources of flora and fauna around the world. Moreover, it contains a large number of high-quality natural images for each species, ranging from dozens to thousands each. Despite a rich source of image data for building identity models, not all Vietnamese plant species appear in this encyclopedia.

By crawling from the mentioned websites, the crawled data was manually cleaned by removing duplicates and irrelevant images. The Vietnamese plant image training dataset consists of 28,046 environmental images from 109 plant species in Vietnam. It has a total of 109 classes, corresponding to 109 species, with an average of 257 images each lass. After training, different deep learning models have been tested on a test dataset of 1071 images. Table I indicates the detailed statistical data of the training dataset, and some samples are illustrated in Fig.8.

TABLE I. STATISTICAL DATA OF THE TRAINING DATASET

| Total amount | Average | Median | Max | Min |
|---|---|---|---|---|
| 28046 | 257 | 190 | 2420 | 86 |

### B. Experiments

In this research, the purpose is to propose a convolutional neural network system to perform feature extraction using different deep learning models in large-scale plant classification methods. The system was conducted in Keras with TensorFlow backend on a computer equipped with CPU Intel Core (™) i5 processor, 8GB RAM, and GPU GTX1070Ti.

Firstly, deep learning models were built to classify plant species through embedding matrices. There are many ways to distinguish and dissect the characteristic matrix, and in this paper, four different deep learning models were used, including MobileNetV2, VGG16, ResnetV2, and Inception Resnet V2. Then, these models were trained using the triplet loss function [35], which used three inputs called anchors, positives, and negatives. The positives are images that have the same class as the anchors, and otherwise, the negatives have a different class. Secondly, the classes were classified with a separated matrix by Support Vector Machine classifier.

The project aims to evaluate the accuracy of different deep convolutional features. It can be seen that due to the overfitting issue, VGG16 has relatively low efficiency. While the evaluation time of the InceptionResnetV2 is much longer than other models, it still achieves average accuracy. ResnetV2 also performs a relatively low loss with its unique structure. And finally, the MobilenetV2 outperforms other models in plant recognition in the SVM classification method. This result proves that MobilenetV2 is a very suitable model for mobile uses as its compactness while running on online servers. For general evaluation, the results are illustrated in Table II, which shows that the MobilenetV2 reached 83.2% accuracy in the best case.

**TABLE II. COMPARISON OF FOUR DIFFERENT DEEP LEARNING MODELS**

| Deep Learning Model | Accuracy on 50 classes (%) | Accuracy on 100 classes (%) | Accuracy on 200 classes (%) | Execution time (seconds) |
|---|---|---|---|---|
| Mobilenet V2 | 75.2 | 79.8 | 83.9 | 8.92 |
| VGG16 | 60.3 | 69.1 | 70.6 | 9.12 |
| Resnet V2 | 69.0 | 70.2 | 74.6 | 19.20 |
| InceptionResnet V2 | 80.1 | 77.2 | 81.2 | 40.21 |

## IV. CONCLUSION

A new dataset on Vietnamese plants has been made in order to evaluate the performance of deep convolutional feature extraction methods in plant identification. In this paper, Resnet50V2, InceptionResnetV2, MobilenetV2, and VGG16 were conducted on this dataset with Support Vector Machine classifier to classify 109 different plant species into their accurate categories. The highest accuracy of 83.9% was achieved from the MobilenetV2 model. The excellent performance proves the efficiency of MobilenetV2 in applying in the mobile applications of plant identification systems as it not only performs high accuracy but also has compactness in the application process. Therefore, a mobile computer system would be developed to help the local citizens to enhance their plant taxonomy knowledge and botanists in species identification techniques. For future research, with an attempt to attain higher accuracy, more classification models would be investigated rather than SVM. A comparison among state-of-the-art classification models would be crucial to upgrade the current work. Future work gears towards using high-performance computing facilities to research a higher performance of plant image identification systems in the natural environment.

## REFERENCES

[1] N. Kumar et al., "*Leaf snap: A Computer vision system for automatic plant species identification*", Proc. Computer Vision – ECCV 2012, Springer, Berlin, Heidelberg, 2012, pp. 502–516.

[2] Du, M., Zhang, S. and Wang, H., "*Supervised Isomap for Plant Leaf Image Classification*", 5th International Conference on Emerging Intelligent Computing Technology and Applications, pp. 627-634, 2009

[3] Hossain, J. and Amin, M.A., "*Leaf Shape Identification Based Plant Biometrics*", 13th International Conference on Computer and Information Technology, Dhaka, Bangladesh, pp. 458-463, 2010.

[4] Du, J.X., Wang, X.F. and Zhang, G.J., "*Leaf shape-based plant species recognition*", Applied Mathematics and Computation, 2007.

[5] Y. Nam and E. Hwang, "*A representation and matching method for shape-based leaf image retrieval*", Journal of KIISE: Software and Applications, vol. 32, no. 11, pp. 1013-1021, 2005.

[6] Y. Nam, J. Park, E. Hwang, and D. Kim, "*Shape-based leaf image retrieval using venation feature*", Proceedings of 2006 Korea Computer Congress, vol. 33, no. 1D, pp. 346-348, 2006.

[7] A. Aakif and M. F. Khan, "*Automatic classification of plants based on their leaves*", Biosystems Engineering, vol. 139, pp. 66–75, 2015.

[8] Wang-Su Jeon and Sang-Yong Rhee, "*Plant Leaf Recognition Using a Convolution Neural Network*", International Journal of Fuzzy Logic and Intelligent Systems, Vol. 17, No. 1, pp. 26-34, March 2017.

[9] Carranza-Rojas, J. and Mata-Montero, E., "*Combining Leaf Shape and Texture for Costa Rican Plant Species Identification*", CLEI Electronic Journal, 19(1), pp. 7, 2016.

[10] G. L. Grinblat, L. C. Uzal, M. G. Larese, and P. M. Granitto, "*Deep learning for plant identification using vein morphological patterns*", Computers and Electronics in Agriculture, pp. 418–424, 2016.

[11] Y. Sun, Y. Liu, G. Wang, and H. Zhang, "*Deep learning for plant identification in natural environment*", Computer Intel Neurosis, 2017.

[12] Du, J.X., Wang, X.F. and Zhang, G.J., "*Leaf shape-based plant species recognition*", Applied Mathematics and Computation, 2007.

[13] Du, M., Zhang, S. and Wang, H., "*Supervised Isomap for Plant Leaf Image Classification*", 5th International Conference on Emerging Intelligent Computing Technology and Applications, Ulsan, South Korea, pp. 627-634, 2009.

[14] Du, J.X., Zhai, C.M. and Wang, Q.P., "*Recognition of plant leaf image based on fractal dimension features*", Neurocomputing, 2013.

[15] Yu Sun, Yuan Liu, Guan Wang, Haiyan Zhang, "*Deep Learning for Plant Identification in Natural Environment*", Computational Intelligence and Neuroscience, May 2017.

[16] Mark Sandler, Andrew Howard, Menglong Zhu, Andrey Zhmoginov, Liang-Chieh Chen, "*MobileNetV2: Inverted Residuals and Linear Bottlenecks*", The IEEE Conference on Computer Vision and Pattern Recognition (CVPR), 2018, pp. 4510-4520.

[17] Mark S., and Andrew H., "*MobileNetV2: The Next Generation of On-Device Computer Vision Networks*", Google Research, April 2018.

[18] Karen Simonyan, Andrew Zisserman, "*Very Deep Convolutional Networks for Large-Scale Image Recognition*", ICLR, Apr 2015.

[19] Muneeb ul H., VGG16 – Convolutional Network for Classification and Detection, Neurohive, Nov 2018.

[20] Nair V., Hinton G.E., "*Rectified linear units improve restricted boltzmann machines*", ICML, 2010.

[21] Lin T.Y. et al., Microsoft COCO: Common objects in context, ECCV, 2014.

[22] Yu S., Yuan L., Guan W., and Haiyan Z., "*Deep Learning for Plant Identification in Natural Environment*", Computational Intelligence and Neuroscience, May 2017.

[23] He K., Zhang X., Ren S., and Sun J., "*Deep residual learning for image recognition*", CVPR, 2016.

[24] Packt, ResNet v2, "*Big Data and Business Intelligence*", Available: https://subscription.packtpub.com/, 2020.

[25] Kaiming He, Xiangyu Zhang, Shaoqing Ren, Jian Sun, "*Deep Residual Learning for Image Recognition*", Microsoft Research, Dec 2015.

[26] Kaiming He, Xiangyu Zhang, Shaoqing Ren, Jian Sun, "*Identity Mappings in Deep Residual Networks*", Microsoft Research, Jul 2016.

[27] Sophia C., Kiat J.H., Teck W.C., MD Abdullah A. M., and Yang L.C., "*Plant Identification on Amazonian and Guiana Shield Flora: NEUON*", in press, Life CLEF 2019 Plant, 2019.

[28] Christian Szegedy, Sergey Ioffe, Vincent Vanhoucke, Alex Alemi, Inception-v4, "*Inception-ResNet and the Impact of Residual Connections on Learning*", Microsoft Research, Aug 2016.
[29] Sandler, M., Howard, A., Zhu, M., and A. Chen L.C., "*Mobilenetv2: Inverted residuals and linear bottlenecks, in press*", IEEE Conference on Computer Vision and Pattern Recognition, Salt Lake City, UT, USA, pp. 4510–4520, June 2018.
[30] Kaiming H., Xiangyu Z., Shaoqing R., and Jian S., "*Deep Residual Learning for Image Recognition*", Microsoft Research, Dec 2015.
[31] Ribeiro B., "*Support vector machines for quality monitoring in a plastic injection molding process*", IEEE Transactions on Systems, Man and Cybernetics C., pp. 401–410, 2005.
[32] Song Q, Hu W, and Xie W., "*Robust support vector machine with bullet hole image classification*", IEEE Transactions on Systems, Man and Cybernetics C., pp. 440–448, 2002.
[33] Phung My Trung et al., "*Encyclopedia of Vietnamese Creatures*", Available: vncreatures.net.
[34] Smithsonian Institution's National Museum of Natural History, Encyclopedia of Life, Available: https://eol.org/.
[35] Marc-Olivier Arsenault, Lossless Triplet loss, Towards Data Science, Feb 2018, Available: https://towardsdatascience.com/lossless-triplet-loss-7e932f990b24.